# Effective Dimensions of Hierarchical Latent Class Models

**Nevin L. Zhang**                                              LZHANG@CS.UST.HK
*Department of Computer Science*
*Hong Kong University of Science and Technology, China*

**Tomáš Kočka**                                                 KOCKA@LISP.VSE.CZ
*Laboratory for Intelligent Systems Prague*
*Prague University of Economics, Czech Republic*

## Abstract

Hierarchical latent class (HLC) models are tree-structured Bayesian networks where leaf nodes are observed while internal nodes are latent. There are no theoretically well justified model selection criteria for HLC models in particular and Bayesian networks with latent nodes in general. Nonetheless, empirical studies suggest that the BIC score is a reasonable criterion to use in practice for learning HLC models. Empirical studies also suggest that sometimes model selection can be improved if standard model dimension is replaced with effective model dimension in the penalty term of the BIC score.

Effective dimensions are difficult to compute. In this paper, we prove a theorem that relates the effective dimension of an HLC model to the effective dimensions of a number of latent class models. The theorem makes it computationally feasible to compute the effective dimensions of large HLC models. The theorem can also be used to compute the effective dimensions of general tree models.

## 1. Introduction

Hierarchical latent class (HLC) models (Zhang, 2002) are tree-structured Bayesian networks (BNs) where leaf nodes are observed while internal nodes are latent. They generalize latent class models (Lazarsfeld and Henry, 1968) and were first identified as a potentially useful class of Bayesian networks by Pearl (1988). We are concerned with learning HLC models from data. A fundamental question is how to select among competing models.

The BIC score (Schwarz, 1978) is a popular metric that researchers use to select among Bayesian network models. It consists of a loglikelihood term that measures the fitness to data and a penalty term that depends linearly upon *standard model dimension*, i.e. the number of linearly independent standard model parameters. When all variables are observed, the BIC score is an asymptotic approximation of (the logarithm) of the marginal likelihood (Schwarz, 1978). It is also *consistent* in the sense that, given sufficient data, the BIC score of the generative model — the model from which data were sampled — is larger than those of any other models that are not equivalent to the generative model.

When latent variables are present, the BIC score is no longer an asymptotic approximation of the marginal likelihood (Geiger *et al.*, 1996). This can be remedied, to some extent, using the concept of effective model dimension. In fact if we replace standard model dimension with effective model dimension in the BIC score, the resulting scoring function, called the *BICe score*, is an asymptotic approximation of the marginal likelihood almost everywhere except for some singular points (Rusakov and Geiger, 2002).

Neither BIC nor BICe have been proved to be consistent for latent variable models. As a matter of fact, it has not even been defined what it means for a model selection criterion to be consistent for latent variable models. Empirical studies suggest that the BIC score is well-behaved in practice for the task of learning HLC models. There are three related search-based algorithms for learning HLC models, namely double hill-climbing (DHC) (Zhang, 2002), single hill-climbing (SHC) (Zhang *et al.*, 2003), and heuristic SHC (HSHC) (Zhang, 2003). In the absence of a theoretically well justified model selection criterion, Zhang (2002) tested DHC with four existing scoring functions, namely the AIC score (Akaike, 1974), the BIC score, the Cheeseman-Stutz (CS) score (Cheeseman and Stutz, 1995), and the holdout logarithmic score (HLS)(Cowell *et al.*, 1999). Both real-world and synthetic data were used. On the real-world data, BIC and CS have enabled DHC to find models that are regarded as the best by domain experts. On the synthetic data, BIC and CS have enabled DHC to find models that either are identical to or resemble closely the true generative models. When coupled with AIC and HLS, on the other hand, DHC performed significantly worse. SHC and HSHC were tested on synthetic data sampled from fairly large HLC models (as much as 28 nodes). Only BIC was used in those tests. In all cases, BIC has enabled SHC and HSHC to find models that either are identical to or resemble closely the true generative models. Those empirical results not only indicate that the algorithms perform well, but also suggest that the BIC is a reasonable scoring function to use for learning HLC models.

The experiments also reveal that model selection can sometimes be improved if the BICe score is used instead of the BIC score. We will explain this in detail in Section 3

In order to use the BICe score in practice, we need a way to compute effective dimensions. This is not a trivial task. The effective dimension of an HLC model is the rank of the Jacobian matrix of the mapping from the parameters of the model to the parameters of the joint distribution of the observed variables. The number of rows in the Jacobian matrix increases exponentially with the number of observed variables. The construction of the Jacobian matrix and the calculation of its rank are both computationally demanding. Moreover they have to be done algebraically or with very high numerical precision to avoid degenerate cases. The necessary precision grows with the size of the matrix.

Settimi and Smith (1998, 1999) studied effective dimensions for two classes of models: trees with binary variables and latent class (LC) models with two observed variables. They have obtained a complete characterization of these two classes. Geiger *et al.* (1996) computed the effective dimensions of a number of models. They conjectured that it is rare for the effective and standard dimensions of an LC model to differ. As a matter of fact, they found only one such model. Kocka and Zhang (2002) found quite a number of LC models whose effective and standard dimensions differ. They also proposed an easily computable formula for estimating effective dimensions of LC models. The estimation formula has been empirically shown to be very accurate.

In this paper, we prove a theorem that relates the effective dimension of an HLC model to the effective dimensions of two other HLC models that contain fewer latent variables. Repeated application of the theorem allows one to reduce the task of computing the effective dimension of an HLC model to subtasks of computing effective dimensions of LC models. This makes it computationally feasible to compute the effective dimensions of large HLC models.

We start in Section 2 with a formal definition of effective dimensions for Bayesian networks with latent variables. In Section 3, we provide empirical evidence that suggest the use of BICe instead of BIC sometimes improves model selection. Section 4 presents the main theorem and Section 5 is devoted to the proof of the theorem. In Section 6, we prove a theorem about effective dimensions of general tree models and explain how this and our main theorem allows one to compute the effective dimension of arbitrary tree models. Finally, concluding remarks are provided in Section 7.

## 2. Effective Dimensions of Bayesian Networks

In this paper, we use capital letters such as $X$ and $Y$ to denote variables and lower case letters such as $x$ and $y$ to denote states of variables. The domain and cardinality of a variable $X$ will be denoted by $\Omega_X$ and $|X|$ respectively. Bold face capital letters such as $\mathbf{Y}$ denote sets of variables. $\Omega_{\mathbf{Y}}$ denotes the Cartesian product of the domains of all variables in the set $\mathbf{Y}$. Elements of $\Omega_{\mathbf{Y}}$ will be denoted by bold lower case letters such as $\mathbf{y}$ and will sometimes be referred to as states of $\mathbf{Y}$. We will consider only variables that have a finite number of states.

Consider a Bayesian network model $M$ that possibly contains latent variables. The *standard dimension* $ds(M)$ of $M$ is the number of linearly independent parameters in the standard parameterization of $M$. The parameters denote, for each variable and each parent configuration of the variable, the probability that the variable is in some state (except one) given the parent configuration. Suppose $M$ consist of $k$ variables $x_1, x_2, \ldots, x_k$. Let $r_i$ and $q_i$ be respectively the number of states of $x_i$ and the number of all possible combinations of the states of its parents. If $x_i$ has no parent, let $q_i$ be 1. Then $ds(M)$ is given by

$$ds(M) = \sum_{i=1}^{k} q_i(r_i - 1).$$

For notational simplicity, denote the standard dimension of $M$ by $n$. Let $\vec{\theta} = (\theta_1, \theta_2, \ldots, \theta_n)$ be a vector of $n$ linearly independent model parameters of $M$. Further let $\mathbf{Y}$ be the set of observed variables. Suppose $\mathbf{Y}$ has $m+1$ possible states. We enumerate the first $m$ states as $\mathbf{y}_1, \mathbf{y}_1, \ldots, \mathbf{y}_m$.

For any $i$ $(1 \leq i \leq m)$, $P(\mathbf{y}_i)$ is a function of the parameters $\vec{\theta}$. So we have a mapping from the $n$ dimensional parameter space (a subspace of $R^n$) to $R^m$, namely $T : (\theta_1, \theta_2, \ldots, \theta_n) \vdash (P(\mathbf{y}_1), P(\mathbf{y}_2), \ldots, P(\mathbf{y}_m))$. The Jacobian matrix of this mapping is the following $m \times n$ matrix:

$$J_M(\vec{\theta}) = [J_{ij}] = [\frac{\partial P(\mathbf{y}_i)}{\partial \theta_j}]$$

For convenience, we will often write the matrix as $J_M = [\frac{\partial P(\mathbf{Y})}{\partial \theta_j}]$, with the understanding that elements of the $j$-th column are obtained by allowing $\mathbf{Y}$ run over all its possible states except one.

For each $i$, $P(\mathbf{y}_i)$ is a function of $\vec{\theta}$. For most commonly used parameterizations of Bayesian networks, it is actually a polynomial function of $\vec{\theta}$. Hence we make the following assumption:

**Assumption 1** *The Bayesian network $M$ is so parameterized that the parameters for the joint distribution of the observed variables are polynomial functions of the parameters for $M$.*

An obvious consequence of the assumption is that elements of $J_M$ are also polynomial functions of $\vec{\theta}$.

For a given value of $\vec{\theta}$, $J_M$ is a matrix of real numbers. Due to Assumption 1, the rank of this matrix is some constant $d$ almost everywhere in the parameter space (Geiger *et al.*, 1996. Also see Section 5.1.). To be more specific, the rank is $d$ everywhere except in a set of measure zero where it is smaller than $d$. The constant is called the *regular rank* of $J_M$.

The regular rank of $J_M$ is also called the *effective dimension* of the Bayesian network model $M$. Hence we denote it by $de(M)$. To understand the term "effective dimension", consider the subspace of $R^m$ spanned by the joint probability $P(\mathbf{Y})$ of observed variables, or equivalently the range of the mapping $T$. The term reflects the fact that, for almost every value of $\vec{\theta}$, a small enough open ball around $T(\vec{\theta})$ resembles Euclidean space of dimension $d$ (Geiger *et al.*, 1996).

There are multiple ways to parameterize a given Bayesian network model. However, the choice of parameterization does not affect the space spanned by the joint probability $P(\mathbf{Y})$. Together with the interpretation of the previous paragraph, this implies that the definition of effective dimension does not depend on the particular parameterization that one uses.

## 3. Selecting among HLC Models

A *hierarchical latent class (HLC) model* is a Bayesian network where (1) the network structure is a rooted tree and (2) the variables at the leaf nodes are observed and all the other variables are not. The observed variables are sometimes referred to as *manifest variables* and all the other variables as *latent variables*. Figure 1 shows the structures of two HLC models. A *latent class (LC) model* is an HLC model where there is only one latent variable.

The theme of this paper is the computation of effective dimensions of HLC models. As mentioned in the introduction, this is interesting because effective dimension, when used in the BIC score, gives us a better approximation of the marginal likelihood. In this section, we give an example to illustrate that the use of effective dimension sometimes also leads to better model selection. We will also motivate and introduce the concept of regularity that will be used in subsequent sections.

### 3.1 An Example of Model Selection

Consider the two HLC models shown in Figure 1. In one experiment, we instantiated the parameters of $M_1$ in a random fashion and sampled a set $D_1$ of 10,000 data records on the observed variables. Then we ran SHC and HSHC on the data set $D_1$ under the guidance of the BIC score. Both algorithms produced model $M_2$. In the following, we explain why, based on $D_1$, one would prefer $M_2$ over $M_1$ if BIC is used for model selection and why $M_1$ would be preferred if BICe is used instead. We argue that $M_1$ should be preferred based on $D_1$ and hence BICe is a better scoring metric for this case.

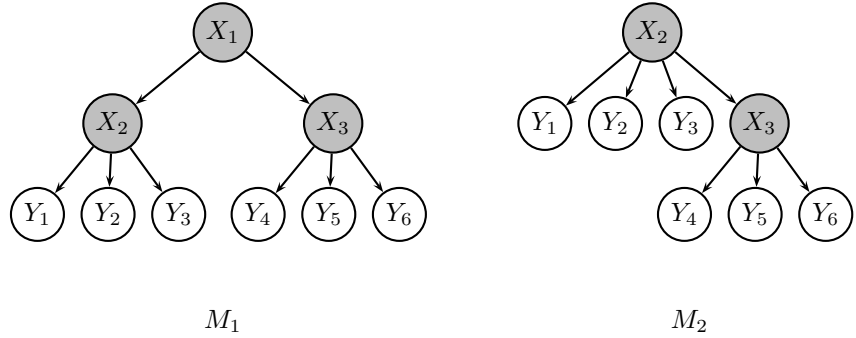

$$M_1 \qquad\qquad M_2$$

Figure 1: Two HLC models. The shaded variables are latent, while the other variables are observed. The cardinality of $X_1$ is 2, while cardinalities of all other variables are 3.

The BIC and BICe scores of a model $M$ given a data set $D$ are defined as follows:

$$BIC(M|D) = logP(D|M,\vec{\theta}^*) - \frac{ds(M)}{2}logN,$$

$$BICe(M|D) = logP(D|M,\vec{\theta}^*) - \frac{de(M)}{2}logN$$

where $\vec{\theta}^*$ is the maximum likelihood estimate of the parameters of $M$ based on $D$ and $N$ is the sample size.

In our example, notice that $M_2$ *includes* $M_1$ in the sense that $M_2$ can represent any probability distributions of the observed variables that $M_1$ can. In fact, if we make the conditional probability distributions of the observed variables in $M_2$ the same as in $M_1$ and set $P_{M_2}(X_2)$ and $P_{M_2}(X_3|X_2)$ such that

$$P_{M_2}(X_2)P_{M_2}(X_3|X_2) = \sum_{X_1} P_{M_1}(X_1)P_{M_1}(X_2|X_1)P_{M_1}(X_3|X_1),$$

then the probability distribution of the observed variables in the two models are identical.

Because $M_2$ includes $M_1$, we have $logP(D_1|M_1,\vec{\theta}_1^*) \leq logP(D_1|M_2,\vec{\theta}_2^*)$. Together with the fact that $D_1$ is sampled from $M_1$, this implies that $logP(D_1|M_1,\vec{\theta}_1^*) \approx logP(D_1|M_2,\vec{\theta}_2^*)$ for sufficiently large enough sample size. The standard dimension of $M_1$ is 45, while that of $M_2$ is 44. Hence

$$BIC(M_1|D_1) < BIC(M_2|D_1).$$

On the other hand, the effective dimensions of $M_1$ and $M_2$ are 43 and 44 respectively. Hence

$$BICe(M_1|D_1) > BICe(M_2|D_1).$$

Model $M_2$ includes $M_1$. The opposite is clearly not true because the effective dimension of $M_1$ is smaller than that of $M_2$. So, $M_2$ is in reality a more complex model than $M_1$. Both model fit data $D_1$ equally well. Hence the simpler one, i.e. $M_1$, should be preferred over the other. This agrees with the choice of the BICe score, while disagrees with the choice of the BIC score. Hence, BICe is more appropriate than BIC in this case.

## 3.2 Regularity

Now consider another model $M_1'$ that is the same as $M_1$ except that the cardinality of $X_1$ is increased from 2 to 3. It is easy to show that $M_2$ includes $M_1'$ and vice versa. So, the two models are equivalent in terms of their capabilities of representing probability distributions of the observed variables. They are hence said to be *marginally equivalent*. However, $M_1'$ has more standard parameters than $M_2$ and hence we would always prefer $M_2$ over $M_1'$. To formalize this consideration, we introduce a concept of regularity.

For a latent variable $Z$ in an HLC model, enumerate its neighbors (parent and children) as $X_1$, $X_2$, ..., $X_k$. An HLC model is *regular* if for any latent variable $Z$,

$$|Z| \leq \frac{\prod_{i=1}^{k} |X_i|}{\max_{i=1}^{k} |X_i|}, \tag{1}$$

and the strict inequality holds when $Z$ has two neighbors and at least one of them is a latent node. Models $M_1$ and $M_2$ are regular, while model $M_1'$ is not.

For any irregular model $M$ there always exists a regular model that is marginally equivalent to $M$ and has fewer standard parameters (Zhang, 2003b). The regular model can be obtained from $M$ as follows: For any latent node that has only two neighbors and its cardinality is no smaller than that of one of the neighbors, then remove the latent node and connect the two neighbors. For any latent node that has more than two neighbors and that violates (1), reduce it's cardinality to the quantity on the right hand side. Repeat both steps until no more changes can be made.

It is also interesting to note that the collection of all regular HLC models for a given set of observed variables is finite (Zhang, 2002). This provides a finite search space for the task of learning regular HLC models.[1] In the rest of this paper, we will consider only regular HLC models.

Before ending this subsection, we point out a nice property of effective model dimension in relation to model inclusion. If an HLC model includes another model, then its effective dimension is no less than that of the latter. As a consequence, two marginally equivalent models have the same effective dimensions and hence the same BICe score. The same is not true for standard model dimension and the BIC score.

## 3.3 The CS and CSe Scores

We have argued on empirical grounds that the BIC score is a reasonable scoring function to use for learning HLC models and that the BICe score can sometimes improve model selection. But the two scores are not free of problems. One problem is that their derivation as Laplace approximations of the marginal likelihood are not valid at the boundary of the parameter space. The CS score in a way alleviates this problem. It involves the BIC score based on completed data and the BIC score based on original data. In other words, it involves two Laplace approximations of the marginal likelihood. It lets errors in the two approximation cancel each other.

Chickering and Heckerman (1997) empirically found the CS score to be a quite accurate approximation of the marginal likelihood and robust at the boundary of the parameter space. They realized the need for effective model dimension in the CS score, although they did not actually use it. This would not have made any differences to their experiments because, for the models they used, the standard and effective dimensions agree.

---

1. The definition of regularity given in this paper is slightly different from the one given in Zhang (2002). Nonetheless, the two conclusions mentioned in this paragraph remain true.

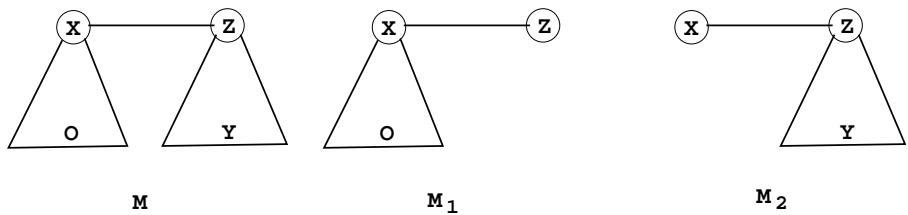

Figure 2: Problem reduction.

We use CSe to refer to the scoring function one obtains by replacing standard model dimension in the CS score with effective model dimensions. Just as BICe is better than BIC as approximations of the marginal likelihood (Geiger *et al.*, 1996), CSe is better than CS. To compute CSe, we also need to calculate effective dimensions.

## 4. Effective Dimensions of HLC Models

As we have seen, effective model dimension is interesting for a number of reasons. Our main result in this paper is a theorem about the effective dimension $de(M)$ of a regular HLC model $M$ that contains more than one latent variable. Let $X$ be the root of $M$, which is a latent node. Because there are at least two latent nodes, there must exist another latent node $Z$ that is a child of $X$. In the following, we will use the terms $X$-*branch* and $Z$-*branch* to respectively refer to the sets of nodes that are separated from $Z$ by $X$ or from $X$ by $Z$. Let $\mathbf{Y}$ be the set of observed variables in the $Z$-branch and let $\mathbf{O}$ be the set of all other observed variables. Note that the $X$-branch doesn't contain the node $X$. The relationship among $X$, $Z$, $\mathbf{Y}$, and $\mathbf{O}$ is depicted in the left-most picture of Figure 2.

The standard parameterization of $M$ includes parameters for $P(X)$ and parameters for $P(Z|X)$. For convenience, we replace those parameters with parameters for $P(X, Z)$. As mentioned at the end of Section 2, such reparameterization does not affect the effective dimension $de(M)$. To reflect the reparameterization, the edge between $X$ and $Z$ is not directed in Figure 2.

Suppose $P(X, Z)$ has $k_0$ parameters $\theta_1^{(0)}, \theta_2^{(0)}, \dots, \theta_{k_0}^{(0)}$. Suppose the conditional distributions of variables in the $X$-branch consists of $k_1$ parameters $\theta_1^{(1)}, \theta_2^{(1)}, \dots, \theta_{k_1}^{(1)}$ and the conditional distributions of variables in the $Z$-branch consists of $k_2$ parameters $\theta_1^{(2)}, \theta_2^{(2)}, \dots, \theta_{k_2}^{(2)}$. For convenience we will sometimes refer to those three groups of parameters using three vectors $\vec{\theta}^{(0)}, \vec{\theta}^{(1)}$ and $\vec{\theta}^{(2)}$ respectively.

In the following, we will define two other HLC models $M_1$ and $M_2$ starting from $M$ and establish a relationship between their effective dimensions and the effective dimension of $M$. In this context, $M$, $M_1$, and $M_2$ are regarded purely as Mathematical objects. The semantics of their variables are of no concern. In particular, a variable $H$ that is latent in $M$ might be designated to be observed in $M_1$ or $M_2$ as part of the definition of those Mathematical objects.

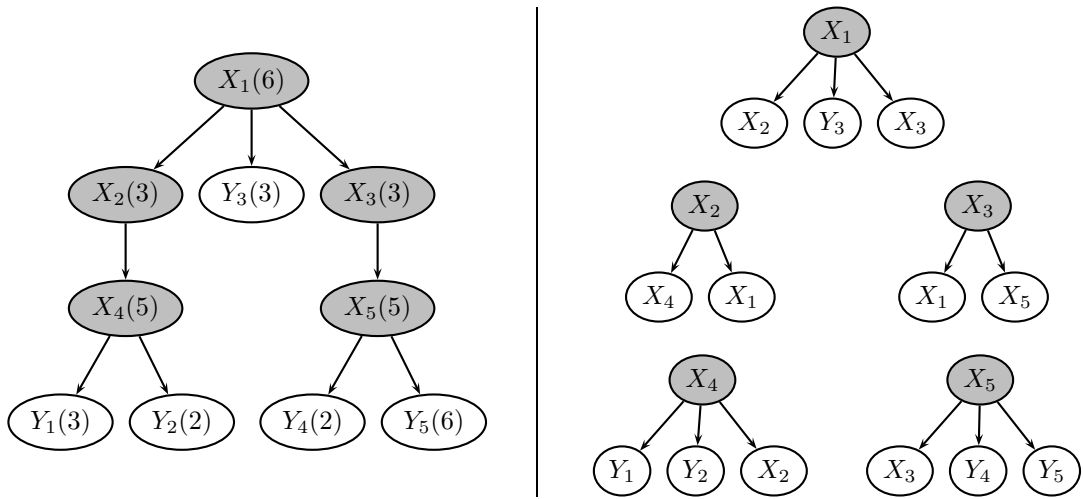

Figure 3: The picture on the left shows an HLC model with five observed and five latent variables, each variable is annotated by its name and its cardinality. The picture on the right shows the components we can decompose the HLC model into by applying Theorem 1. Latent variables are shaded, while observed variables are not.

We obtain a Bayesian network model $B_1$ from $M$ by deleting the $Z$-branch. Strictly speaking $B_1$ is not Bayesian network due to the parameterization it inherits from $M$: instead of probability tables $P(X)$ and $P(Z|X)$, we have table $P(X, Z)$. But $P(X)$ and $P(Z|X)$ can readily be obtained from $P(X, Z)$. With this in mind, we view $B_1$ as a Bayesian network. This network is obviously tree-structured. It's leaf variables include those in the set $\mathbf{O}$ and the variable $Z$. We define $M_1$ to be the HLC model that share the same structure as $B_1$ and where the variable $Z$ and all the variables in $\mathbf{O}$ are observed. The parameters of $M_1$ are $\vec{\theta}^{(0)}$ and $\vec{\theta}^{(1)}$.

Similarly let $B_2$ be the Bayesian network model obtained from $M$ by deleting the $X$-branch. It is a tree-structure and its leaf variables include those in $\mathbf{Y}$ and the variable $X$. We define $M_2$ to be the HLC model that share the same structure as $B_2$ and where the variable $X$ and all the variables in $\mathbf{Y}$ are observed. The parameters of $M_2$ are $\vec{\theta}^{(0)}$ and $\vec{\theta}^{(2)}$.

**Theorem 1** *Suppose $M$ is a regular HLC model that contains two or more latent nodes. Then the two HLC models $M_1$ and $M_2$ defined in the text are also regular. Moreover,*

$$de(M) = de(M_1) + de(M_2) - [ds(M_1) + ds(M_2) - ds(M)]. \tag{2}$$

*In words, the effective dimension of $M$ equals the sum of the effective dimensions of $M_1$ and $M_2$ minus the number of common parameters that $M_1$ and $M_2$ share.*

To appreciate the significance of this theorem, consider the task of computing the effective dimension of a regular HLC model that contains two or more latent nodes. By repeatedly applying the theorem, we can reduce the task into subtasks of calculating effective dimensions of LC models. As an example, consider the HLC model depicted by the picture on the left in Figure 3. Theorem 1 allows us to, for the purpose of computing its effective dimension, decompose the HLC model into five LC models, which are shown on the right in Figure 3.

How might one compute the effective dimension of an LC model? One way is to use the algorithm suggested by Geiger *et al.* (1996). The algorithm first symbolically computes the Jacobian matrix, which is possible due to Assumption 1. Then it randomly assigns values to the parameters, resulting a numerical matrix. The rank of the numerical matrix is computed by diagonalization. Because the rank of Jacobian matrix equals the effective dimension of the LC model almost everywhere, we get the regular rank with probability one. This algorithm has recently been implemented by Rusakov and Geiger (2003). Kocka and Zhang (2002) suggest an alternative algorithm that computes an upper bound. The algorithm is fast and has been empirically shown to produce extremely tight bounds.

Going back to our example, the effective dimension of the LC models for $X_1$, $X_2$, $X_3$, $X_4$ and $X_5$ are 26, 23, 23, 34 and 17 respectively. Thus the effective dimension of the HLC model in Figure 3 is $26+23+34+23+17-(5*3-1)-(3*6-1)-(6*3-1)-(3*5-1) = 61$. In contrast, the standard dimension of the model is $5+6*2+6*2+6*2+3*4+5*5+5+3*4+5*2+5 = 110$.

## 5. Proof of Main Result

This section is devoted to the proof of Theorem 1. We begin with some properties of Jacobian matrices of Bayesian network models.

### 5.1 Properties of Jacobian Matrices

Consider the Jacobian matrix $J_M$ of a Bayesian network model $M$. It is a matrix parameterized by the parameters $\vec{\theta}$ of $M$. Let $v_1$, $v_2$, ..., $v_m$ be column vectors of $J_M$.

**Lemma 1** *A number of column vectors $v_1$, $v_2$, ..., $v_m$ of the Jacobian matrix $J_M$ are either linearly dependent everywhere or linearly independent almost everywhere. They are linearly dependent everywhere if and only if there exists at least one column vector $v_j$ that can be expressed as a linear combination of other column vectors everywhere.*

**Proof**: Consider diagonalizing the following transposed matrix:

$$[v_1, v_2, \ldots, v_m]^T.$$

According to Assumption 1, elements of the matrix are polynomials (of $\vec{\theta}$). Hence we would multiply rows with polynomials or fraction of polynomials. Of course, we need also to add one row to another row. At the end of the process, we get a diagonal matrix whose nonzero elements are polynomials or fractions of polynomials. Suppose there are $k$ nonzero rows and suppose they correspond to $v_1$, $v_2$, ..., $v_k$.

Because elements of the diagonalized matrix are polynomials or fractions of polynomials, they are well-defined [2] and nonzero almost everywhere (i.e. for almost all values of $\vec{\theta}$). If $k=m$, then the $m$ vectors are linearly independent of each other almost everywhere.

---

2. A fraction is not well defined if the denominator is zero.

If $k < m$, there exist, for each $j$ ($k < j \leq m$), polynomials or fractions of polynomials $c_i$ ($1 \leq i \leq k$) such that

$$v_j = \sum_{i=1}^{k} c_i v_i. \tag{3}$$

The coefficients $c_i$'s can be determined by tracing the diagonalization process. So $v_j$ can be expressed as a linear combination of $\{v_i | i = 1, \ldots, k\}$ everywhere [3]. $\square$

Although it might sound trivial, this lemma is actually quite interesting. This is because $J_M$ is a parameterized matrix. The first part, for example, implies that there do not exist two subspaces of the parameter space that both have nonzero measures such that the $m$ vectors are linearly independent in one subspace while linearly dependent in the other.

If $m$ is the total number of column vectors of $J_M$, we get the following lemma:

**Lemma 2** *In the Jacobian matrix $J_M$, there exists a collection of column vectors that form a basis of its column space almost everywhere. The number of vectors in the collection equals to the regular rank of the matrix. Moreover, the collection can be chosen to include any given set of column vectors that are linearly independent almost everywhere.*

**Proof**: The first part has already been proved. The second part follows from the definition of regular rank. The last part is true because we could start the diagonalization process with the transpose of the vectors in the set on the top of the matrix. $\square$

## 5.2 Proof of Theorem 1

We now set out to prove Theorem 1. It is straightforward to verify that the HLC models $M_1$ and $M_2$ are regular. So it suffices to prove equation (2). This is what we do in the rest of this section.

The set of observed variables in $M$ is $\mathbf{O} \cup \mathbf{Y}$, the set of observed variables in $M_1$ is $\mathbf{O} \cup \{Z\}$ and the set of observed variables in $M_2$ is $\mathbf{Y} \cup \{X\}$. Hence the Jacobian matrices of models $M$, $M_1$, and $M_2$ can be respectively written as follows:

$$J_M = [\frac{\partial P(\mathbf{O}, \mathbf{Y})}{\partial \theta_1^{(0)}}, \ldots, \frac{\partial P(\mathbf{O}, \mathbf{Y})}{\partial \theta_{k_0}^{(0)}}; \frac{\partial P(\mathbf{O}, \mathbf{Y})}{\partial \theta_1^{(1)}}, \ldots, \frac{\partial P(\mathbf{O}, \mathbf{Y})}{\partial \theta_{k_1}^{(1)}}; \frac{\partial P(\mathbf{O}, \mathbf{Y})}{\partial \theta_1^{(2)}}, \ldots, \frac{\partial P(\mathbf{O}, \mathbf{Y})}{\partial \theta_{k_2}^{(2)}}]$$

$$J_{M_1} = [\frac{\partial P(\mathbf{O}, Z)}{\partial \theta_1^{(0)}}, \ldots, \frac{\partial P(\mathbf{O}, Z)}{\partial \theta_{k_0}^{(0)}}; \frac{\partial P(\mathbf{O}, Z)}{\partial \theta_1^{(1)}}, \ldots, \frac{\partial P(\mathbf{O}, Z)}{\partial \theta_{k_1}^{(1)}}]$$

$$J_{M_2} = [\frac{\partial P(X, \mathbf{Y})}{\partial \theta_1^{(0)}}, \ldots, \frac{\partial P(X, \mathbf{Y})}{\partial \theta_{k_0}^{(0)}}; \frac{\partial P(X, \mathbf{Y})}{\partial \theta_1^{(2)}}, \ldots, \frac{\partial P(X, \mathbf{Y})}{\partial \theta_{k_2}^{(2)}}]$$

---

3. There is a subtle point here. Being fractions of polynomials of $\vec{\theta}$, the $c_i$'s might be undefined for some values of $\vec{\theta}$. So from equation (3) alone, we cannot conclude that $v_j$ linearly depends on $\{v_i | i = 1, \ldots, k\}$ everywhere.

The conclusion is nonetheless true for two reasons. First the set of $\vec{\theta}$ values where the $c_i$'s are undefined has measure zero. Second, if $v_j$ does not linearly depend on $\{v_i | i = 1, \ldots, k\}$ at one value of $\vec{\theta}$, then the same would be true in a sufficiently small and nonetheless measure-positive ball around that value.

It is clear that there is a one-to-one correspondence between the first $k_0+k_1$ column vectors of $J_M$ with the column vectors of $J_{M_1}$ and there is a one-to-one correspondence between the first $k_0$ and the last $k_2$ column vectors of $J_M$ with the column vectors of $J_{M_2}$. We will first show

**Claim 1**: The first $k_0$ vectors of $J_M$ ($J_{M_1}$ or $J_{M_2}$) are linearly independent almost everywhere.

Together with Lemma 2, Claim 1 implies that there is a collection of column vectors in $J_{M_1}$ that includes the first $k_0$ vectors and that is a basis of the column space of $J_{M_1}$ almost everywhere. In particular, this implies that $de(M_1) \geq k_0$. Suppose $de(M_1)=k_0+r$. Without loss of generality, suppose the basis vectors are

$$\frac{\partial P(\mathbf{O}, Z)}{\partial \theta_1^{(0)}}, \ldots, \frac{\partial P(\mathbf{O}, Z)}{\partial \theta_{k_0}^{(0)}}; \frac{\partial P(\mathbf{O}, Z)}{\partial \theta_1^{(1)}}, \ldots, \frac{\partial P(\mathbf{O}, Z)}{\partial \theta_r^{(1)}}. \tag{4}$$

By symmetry, we can assume that $de(M_2)=k_0+s$ where $s \geq 0$ and that the following column vectors form a basis for $J_{M_2}$ almost everywhere:

$$\frac{\partial P(X, \mathbf{Y})}{\partial \theta_1^{(0)}}, \ldots, \frac{\partial P(X, \mathbf{Y})}{\partial \theta_{k_0}^{(0)}}; \frac{\partial P(X, \mathbf{Y})}{\partial \theta_1^{(2)}}, \ldots, \frac{\partial P(X, \mathbf{Y})}{\partial \theta_s^{(2)}}. \tag{5}$$

Now consider the following list of vectors in $J_M$:

$$\frac{\partial P(\mathbf{O}, \mathbf{Y})}{\partial \theta_1^{(0)}}, \ldots, \frac{\partial P(\mathbf{O}, \mathbf{Y})}{\partial \theta_{k_0}^{(0)}}; \frac{\partial P(\mathbf{O}, \mathbf{Y})}{\partial \theta_1^{(1)}}, \ldots, \frac{\partial P(\mathbf{O}, \mathbf{Y})}{\partial \theta_r^{(1)}}; \frac{\partial P(\mathbf{O}, \mathbf{Y})}{\partial \theta_1^{(2)}}, \ldots, \frac{\partial P(\mathbf{O}, \mathbf{Y})}{\partial \theta_s^{(2)}}. \tag{6}$$

We will show

**Claim 2**: All column vectors of $J_M$ linearly depend on the vectors listed in (6) everywhere.

**Claim 3**: The vectors listed in (6) are linearly independent almost everywhere.

Those two claims imply that the vectors listed in (6) form a basis of the column space of $J_M$ almost everywhere. Therefore

$$de(M) = k_0+r+s = de(M_1)+de(M_2)-k_0.$$

It is clear that $k_0=ds(M_1)+ds(M_2)-ds(M)$. Therefore Theorem 1 is proved. $\square$

## 5.3 Proof of Claim 1

**Lemma 3** *Let $Z$ be a latent node in an HLC model $M$ and $\mathbf{Y}$ be the set of the observed nodes in the subtree rooted at $Z$. If $M$ is regular, then we can set conditional distributions of nodes in the subtree in such a way that they encode an injective mapping $\rho$ from $\Omega_Z$ to $\Omega_{\mathbf{Y}}$ in the sense that $P(\mathbf{Y}=\rho(z)|Z=z)=1$ for all $z \in \Omega_Z$.*

**Proof**: We prove this lemma by induction on the number of latent nodes in the subtree rooted at $Z$. First consider the case when there is only one latent node, namely $Z$. In this case, $Z$ is the parent of all nodes in $\mathbf{Y}$. Enumerate all these nodes as $Y_1, Y_2, \ldots, Y_k$. Because $M$ is regular, we have $|Z| \leq \prod_{i=1}^{k} |Y_i|$. Hence we can define an injective mapping $\rho$ from $\Omega_Z$ to $\Omega_{\mathbf{Y}} = \prod_{i=1}^{k} \Omega_{Y_i}$. For each state $z$ of $Z$, $\rho(z)$ can be written as $y = (y_1, y_2, \ldots, y_k)$, where $y_i$ is a state of $Y_i$. Now if we set

$$P(Y_i=y_i|Z=z) = 1,$$

then $P(\mathbf{Y}=\rho(z)|Z=z) = 1$.

Now consider the case when there are at least two hidden nodes in the subtree rooted at $Z$. Let $W$ be one such latent node that has no latent node descendants. Let $\mathbf{Y}^{(1)}$ be the set of observed nodes in the subtree rooted at $W$ and $\mathbf{Y}^{(2)} = \mathbf{Y} \backslash \mathbf{Y}^{(1)}$. By the induction hypothesis, we can parameterize the subtree rooted at $W$ in such a way that it encodes an injective mapping from $\Omega_W$ to $\Omega_{\mathbf{Y}^{(1)}}$. Moreover, if all nodes below $W$ are removed from $M$, $M$ remains a regular HLC model. In that model, we can parameterize the subtree rooted at $Z$ in such a way that it encodes an injective mapping from $\Omega_Z$ to $\Omega_{(W, \mathbf{Y}^{(2)})} = \Omega_W \times \Omega_{\mathbf{Y}^{(2)}}$. Together, those two facts prove the lemma. $\square$

**Corollary 1** *Let $Z$ be a latent node in an HLC model $M$. Suppose $Z$ have a latent neighbor $X$. Let $\mathbf{Y}$ be the set of the observed nodes separated from $X$ by $Z$. If $M$ is regular, then we can set probability distributions of nodes separated from $X$ by $Z$ in such a way that they encode an injective mapping $\rho$ from $\Omega_Z$ to $\Omega_{\mathbf{Y}}$ in the sense that $P(\mathbf{Y}=\rho(z)|Z=z) = 1$ for all $z \in \Omega_Z$.*

**Proof**: The corollary follows readily from Lemma 3 and the property of the root-walking operation (Zhang, 2002). $\square$

**Proof of Claim 1**: Consider the following matrix

$$[\frac{\partial P(X, Z)}{\partial \theta_1^{(0)}} \ldots, \frac{\partial P(X, Z)}{\partial \theta_{k_0}^{(0)}}] \tag{7}$$

Because $\theta_1^{(0)}, \theta_2^{(0)}, \ldots, \theta_{k_0}^{(0)}$ are the parameters for the joint distribution $P(X, Z)$, this matrix is the identity matrix if the rows are properly arranged. So its column vectors are linearly independent almost everywhere.

Now consider the first $k_0$ column vectors of $J_M$: $\partial P(\mathbf{O}, \mathbf{Y})/\partial \theta_1^{(0)}, \ldots, \partial P(\mathbf{O}, \mathbf{Y})/\partial \theta_{k_0}^{(0)}$. They must be linearly independent almost everywhere. If not, one of the vectors, say $\partial P(\mathbf{O}, \mathbf{Y})/\partial \theta_{k_0}^{(0)}$, would linearly depend on the rest everywhere according to Lemma 1. Observe that for any $i$ ($1 \leq i \leq k_0$),

$$\frac{\partial P(\mathbf{O}, \mathbf{Y})}{\partial \theta_i^{(0)}} \quad = \quad \sum_{X, Z} P(\mathbf{O}|X) P(\mathbf{Y}|Z) \frac{\partial P(X, Z)}{\partial \theta_i^{(0)}}.$$

Choose $P(\mathbf{O}|X)$ and $P(\mathbf{Y}|Z)$ as in Corollary 1. The vector $\partial P(\mathbf{O}, \mathbf{Y})/\partial \theta_i^{(0)}$ might contain zero elements. If we remove the zero elements, what remains of the vector is identical to $\partial P(X, Z)/\partial \theta_i^{(0)}$. So we can conclude that $\partial P(X, Z)/\partial \theta_{k_0}^{(0)}$ linearly depends on $\partial P(X, Z)/\partial \theta_1^{(0)} \ldots, \partial P(X, Z)/\partial \theta_{k_0-1}^{(0)}$ everywhere, which contradicts the conclusion of the previous paragraph. Hence the first $k_0$ vectors of $J_M$ must be linearly independent almost everywhere.

It is evident that, using similar arguments, we can also show that the first $k_0$ vectors of $J_{M_1}$ ($J_{M_2}$) are linearly independent almost everywhere. Claim 1 is therefore proved. $\square$

### 5.4 Proof of Claim 2

Every column vector of $J_{M_1}$ linearly depends on vectors listed in (4) everywhere. Observe that

$$
\begin{aligned}
\frac{\partial P(\mathbf{O}, \mathbf{Y})}{\partial \theta_i^{(0)}} &= \sum_Z P(\mathbf{Y}|Z)\frac{\partial P(\mathbf{O}, Z)}{\partial \theta_i^{(0)}}, i = 1, \ldots, k_0 \\
\frac{\partial P(\mathbf{O}, \mathbf{Y})}{\partial \theta_i^{(1)}} &= \sum_Z P(\mathbf{Y}|Z)\frac{\partial P(\mathbf{O}, Z)}{\partial \theta_i^{(1)}}, i = 1, \ldots, k_1.
\end{aligned}
$$

Therefore every column vector of $J_M$ that corresponds to vectors in $J_{M_1}$ linearly depends on the first $k_0+r$ vectors listed in (6) everywhere.

By symmetry, every column vector of $J_M$ that corresponds to vectors in $J_{M_2}$ linearly depends on the first $k_0$ and the last $s$ vectors listed in (6) everywhere. The claim is proved. $\square$

### 5.5 Proof of Claim 3

We prove this claim by contradiction. Assume the vectors listed in (6) were not linearly independent almost everywhere. According to Lemma 1, one of them, say $v$, must linearly depend on the rest everywhere. Because of Claim 1 and Lemma 2, we can assume that $v$ is among the last $r+s$ vectors. Without loss of generality, we assume that $v$ is $\partial P(\mathbf{O}, \mathbf{Y})/\partial \theta_s^{(2)}$. Then for any value of $\vec{\theta}$, there exist real numbers $c_i$ ($1 \le i \le k_0$), $c_i^{(1)}$ ($1 \le i \le r$), and $c_i^{(2)}$ ($1 \le i \le s-1$) such that

$$
\frac{\partial P(\mathbf{O}, \mathbf{Y})}{\partial \theta_s^{(2)}} = \sum_{i=1}^{k_0} c_i \frac{\partial P(\mathbf{O}, \mathbf{Y})}{\partial \theta_i^{(0)}} + \sum_{i=1}^{r} c_i^{(1)} \frac{\partial P(\mathbf{O}, \mathbf{Y})}{\partial \theta_i^{(1)}} + \sum_{i=1}^{s-1} c_i^{(2)} \frac{\partial P(\mathbf{O}, \mathbf{Y})}{\partial \theta_i^{(2)}}.
$$

Note that in the last term on the right hand side, $i$ runs from 1 to $s-1$.

The parameter vector $\vec{\theta}$ consists of three subvectors $\vec{\theta}^{(0)}$, $\vec{\theta}^{(1)}$ and $\vec{\theta}^{(2)}$. Set the parameters $\vec{\theta}^{(1)}$ (for the $X$-branch) as in Lemma 3. Then there exists an injective mapping $\rho$ from $\Omega_X$ to $\Omega_\mathbf{O}$ such that

$$
P(\mathbf{O}=\rho(x)|X=x) = 1 \text{ for all } x \in \Omega_X. \tag{8}
$$

For each of the vectors in (6), consider the subvector consisting only of elements for those states of $\mathbf{O}$ that are the images of states of $X$ under the mapping $\rho$. Such subvectors will be denoted by $\partial P(\mathbf{O}_X, \mathbf{Y})/\partial \theta_i^{(0)}$, $\partial P(\mathbf{O}_X, \mathbf{Y})/\partial \theta_i^{(1)}$, and $\partial P(\mathbf{O}_X, \mathbf{Y})/\partial \theta_i^{(2)}$. For any values of $\vec{\theta}^{(0)}$ and $\vec{\theta}^{(2)}$, we still have

$$\frac{\partial P(\mathbf{O}_X, \mathbf{Y})}{\partial \theta_s^{(2)}} = \sum_{i=1}^{k_0} c_i \frac{\partial P(\mathbf{O}_X, \mathbf{Y})}{\partial \theta_i^{(0)}} + \sum_{i=1}^{r} c_i^{(1)} \frac{\partial P(\mathbf{O}_X, \mathbf{Y})}{\partial \theta_i^{(1)}} + \sum_{i=1}^{s-1} c_i^{(2)} \frac{\partial P(\mathbf{O}_X, \mathbf{Y})}{\partial \theta_i^{(2)}}. \quad (9)$$

Consider the first two terms on the right hand side:

$$\sum_{i=1}^{k_0} c_i \frac{\partial P(\mathbf{O}_X, \mathbf{Y})}{\partial \theta_i^{(0)}} + \sum_{i=1}^{r} c_i^{(1)} \frac{\partial P(\mathbf{O}_X, \mathbf{Y})}{\partial \theta_i^{(1)}}$$

$$= \sum_{i=1}^{k_0} c_i \sum_Z P(\mathbf{Y}|Z) \frac{\partial P(\mathbf{O}_X, Z)}{\partial \theta_i^{(0)}} + \sum_{i=1}^{r} c_i^{(1)} \sum_Z P(\mathbf{Y}|Z) \frac{\partial P(\mathbf{O}_X, Z)}{\partial \theta_i^{(1)}}$$

$$= \sum_{\mathbf{Z}} P(\mathbf{Y}|Z) \{ \sum_{i=1}^{k_0} c_i \frac{\partial P(\mathbf{O}_X, Z)}{\partial \theta_i^{(0)}} + \sum_{i=1}^{r} c_i^{(1)} \frac{\partial P(\mathbf{O}_X, Z)}{\partial \theta_i^{(1)}} \}$$

Because of (8) and the fact that $P(\mathbf{O}, Z) = \sum_X P(X, Z) P(\mathbf{O}|X)$, the column vector $\partial P(\mathbf{O}_X, Z)/\partial \theta_i^{(0)}$ is identical to the vector $\partial P(X, Z)/\partial \theta_i^{(0)}$. As we have argued when proving Claim 1, the vectors $\{\partial P(X, Z)/\partial \theta_i^{(0)} | i=1, \ldots, k_0\}$ constitute a basis for the $k_0$-dimensional Euclidian space. This implies that, each of the vectors $\partial P(\mathbf{O}_X, Z)/\partial \theta_i^{(1)}$ can be represented as a linear combination of the vectors $\{\partial P(\mathbf{O}_X, Z)/\partial \theta_i^{(0)} | i = 1, \ldots, k_0\}$. Consequently, there exist $c_i'$ $(1 \le i \le k_0)$ such that

$$\sum_{i=1}^{k_0} c_i \frac{\partial P(\mathbf{O}_X, Z)}{\partial \theta_i^{(0)}} + \sum_{i=1}^{r} c_i^{(1)} \frac{\partial P(\mathbf{O}_X, Z)}{\partial \theta_i^{(1)}} = \sum_{i=1}^{k_0} c_i' \frac{\partial P(\mathbf{O}_X, Z)}{\partial \theta_i^{(0)}}$$

Hence

$$\sum_{i=1}^{k_0} c_i \frac{\partial P(\mathbf{O}_X, \mathbf{Y})}{\partial \theta_i^{(0)}} + \sum_{i=1}^{r} c_i^{(1)} \frac{\partial P(\mathbf{O}_X, \mathbf{Y})}{\partial \theta_i^{(1)}} = \sum_{i=1}^{k_0} c_i' \frac{\partial P(\mathbf{O}_X, \mathbf{Y})}{\partial \theta_i^{(0)}}$$

Combining this equation with equation (9), we get

$$\frac{\partial P(\mathbf{O}_X, \mathbf{Y})}{\partial \theta_s^{(2)}} = \sum_{i=1}^{k_0} c_i' \frac{\partial P(\mathbf{O}_X, \mathbf{Y})}{\partial \theta_i^{(0)}} + \sum_{i=1}^{s-1} c_i^{(2)} \frac{\partial P(\mathbf{O}_X, \mathbf{Y})}{\partial \theta_i^{(2)}}.$$

Because of (8) and the fact that the fact that $P(\mathbf{O}, \mathbf{Y}) = \sum_X P(X, \mathbf{Y}) P(\mathbf{O}|X)$, the column vector $\partial P(\mathbf{O}_X, \mathbf{Y})/\partial \theta_i^{(1)}$ is identical to the vector $\partial P(X, \mathbf{Y})/\partial \theta_i^{(1)}$ and the column vector $\partial P(\mathbf{O}_X, \mathbf{Y})/\partial \theta_i^{(2)}$ is identical to the vector $\partial P(X, \mathbf{Y})/\partial \theta_i^{(2)}$. Hence

$$\frac{\partial P(X, \mathbf{Y})}{\partial \theta_s^{(2)}} = \sum_{i=1}^{k_0} c_i' \frac{\partial P(X, \mathbf{Y})}{\partial \theta_i^{(0)}} + \sum_{i=1}^{s-1} c_i^{(2)} \frac{\partial P(X, \mathbf{Y})}{\partial \theta_i^{(2)}}.$$

This contradicts the fact that the vectors in the equation form a basis for the column space of $J_{M_2}$ almost everywhere (see (5) in Section 5.2) Therefore, Claim 3 must be true. □

## 6. Effective Dimensions of Trees

Let us use the term *tree model* to refer to Markov random fields on undirected trees over a finite number of random variables. If we root a tree model at any of its nodes, we get a tree-structured Bayesian network model. In a tree model, define *leaf nodes* be those that have only one neighbor. An HLC model is a tree model where all leaf nodes are observed while all others are latent.

It turns out that Theorem 1 enables us to compute the effective dimension of any tree model. Consider an arbitrary tree model. If some of its leaf nodes are latent, we can remove such nodes without affecting its effective dimension.

After removing latent leaf nodes, all the leaf nodes are observed. If some non-leaf nodes are also observed, we can decompose the model into submodels at any observed non-leaf node. The following theorem tells us how the model and the submodels are related in terms of effective dimensions.

**Theorem 2** *Suppose $Y$ is an observed non-leaf node in a tree model $M$. If $M$ decomposes at $Y$ into $k$ submodels $M_1, \ldots, M_k$, then*

$$de(M) = \sum_{i=1}^{k} de(M_i) - (k-1)(|Y|-1).$$

After all possible decompositions, the final submodels either do not contain latent nodes or are HLC models. Effective dimensions of submodels with no latent variables are simply their standard dimensions. If an HLC submodel is irregular, we make it regular by applying the transformation mentioned at the end of Section 3.2. The transformation does not affect the effective dimensions of the submodels. Finally, effective dimensions of regular HLC submodels can be computed using Theorem 1.

**Proof of Theorem 2**: It is possible to prove this theorem starting from the Jacobian matrix. Here we take a less formal but more revealing approach.

It suffices to consider case of $k$ being 2. The two submodels $M_1$ and $M_2$ share only one node, namely $Y$. Let $\mathbf{O_1}$ and $\mathbf{O_2}$ be respectively the sets of observed nodes in those two submodels excluding $Y$. Root $M$ at $Y$. Then we have

$$P(Y, \mathbf{O_1}, \mathbf{O_2})P(Y) = P(\mathbf{O_1}, Y)P(\mathbf{O_2}, Y).$$

Let $\vec{\theta}_0$ be the set of parameters in the distribution $P(Y)$, $\vec{\theta}_1$ and $\vec{\theta}_2$ be respectively the sets of parameters in the conditional probability distributions of nodes in $M_1$ and $M_2$. Consider fixing $\vec{\theta}_0$ and letting $\vec{\theta}_1$ and $\vec{\theta}_2$ vary. In this case, the space spanned by $P(Y)$ consists of only one vector, namely $\vec{\theta}_0$ itself. Moreover, there is a one-to-one correspondence between vectors in the space spanned by $P(Y, \mathbf{O_1}, \mathbf{O_2})$ and vectors in the Cartesian product of the spaces spanned by $P(\mathbf{O_1}, Y)$ and $P(\mathbf{O_2}, Y)$. Now let $\vec{\theta}_0$ vary. This adds $|Y|-1$ dimensions to each of the four spaces spanned by $P(Y, \mathbf{O_1}, \mathbf{O_2})$, $P(Y)$, $P(\mathbf{O_1}, Y)$, and $P(\mathbf{O_2}, Y)$. Consequently, we have

$$de(M) = de(M_1) + de(M_2) - (|Y|-1).$$

The theorem is proved. □

## 7. Concluding Remarks

In this paper we study the effective dimensions of HLC models. The work is motivated by empirical evidence that the BIC behaves quite well when used with several hill-climbing algorithms for learning HLC models and that the BICe score sometimes leads to better model selection than the BIC score. We have proved a theorem that relates the effective dimension of an HLC model to the effective dimensions of two other HLC models that contain fewer latent variables. Repeated application of the theorem allows one to reduce the task of computing the effective dimension of an HLC model to subtasks of computing effective dimensions of LC models. This makes it computationally feasible to compute the effective dimensions of large HLC models. In addition, we have proved a theorem about effective dimensions of general tree models. This and our main theorem allows one to compute the effective dimension of arbitrary tree models.

### Acknowledgements

This work was initiated when the authors were visiting Department of Computer Science, Aalborg University, Denmark. We thank Poul S. Eriksen, Finn V. Jensen, Jiri Vomlel, Marta Vomlelova, Thomas D. Nielsen, Olav Bangso, Jose Pena, Kristian G. Olesen. We are also grateful to the annonymous reviewers whose comments have helped us greatly in improving this paper. Research on this paper was partially supported by GA CR grant 201/02/1269 and by Hong Kong Research Grant Council under grant HKUST6088/01E.